\pdfoutput=1
\documentclass[11pt]{article}

\usepackage[final]{acl}
 \usepackage{booktabs}
\usepackage{times}
\usepackage{latexsym}
\usepackage{CJKutf8}
\usepackage[utf8]{inputenc}

\usepackage[T1]{fontenc}

\usepackage[utf8]{inputenc}

\usepackage{microtype}

\usepackage{inconsolata}

\usepackage{graphicx}

\usepackage{float}

\title{Stylistic Evolution and LLM Neutrality in Singlish Language}

\author{Linus Tze En Foo \\
  Independent Researcher \\
  Singapore, Singapore \\
  \texttt{linusfoo@outlook.sg} \\
  \And
  Weihan Angela Ng \\
  ETH Zürich \\
  Zürich, Switzerland \\
  \texttt{angela.ng@gess.ethz.ch} \\
  \AND
  Wenkai Li \\
  Carnegie Mellon University \\
  Pittsburgh, United States \\
  \texttt{wenkail@andrew.cmu.edu} \\
  \And
  Lynnette Hui Xian Ng \\
  Carnegie Mellon University \\
  Pittsburgh, United States \\
  \texttt{lynnetteng@cmu.edu} \\
}

\begin{document}
\maketitle
\begin{abstract}
Singlish is a creole rooted in Singapore's multilingual environment 
that continues to evolve alongside social and technological change. 
We examine diachronic stylistic change across a decade of informal 
digital messages and ask whether Large Language Models (LLMs) can 
generate temporally neutral outputs approximating the stable characteristics of the variety. Using lexical, pragmatic, psycholinguistic, and 
encoder-based features, we find that stylistic separability increases 
with temporal distance, driven primarily by structural features such as length and complexity. Evaluated against a null distribution baseline, most LLMs fail to achieve both authenticity and temporal neutrality simultaneously at their most authentic prompting configuration. This reveals a structural trade-off: models 
generating realistic Singlish tend to inherit its temporal biases, while 
temporally neutral models tend to produce inauthentic outputs. These findings 
position temporal neutrality as a diagnostic metric for assessing 
sociolectal grounding in LLMs.

\end{abstract}

\section{Introduction}
Singlish is a creole-like variety of English widely spoken in Singapore. As a contact language, it incorporates grammatical and lexical elements from Chinese, Malay, Tamil, and their dialects, giving rise to a rich and dynamic linguistic system \cite{tan2005english}. Singlish continues to evolve as cultural norms, technologies, and generational usage patterns shift \cite{satibaldieva2024dynamics}. These changes are especially apparent in Singlish due to Singapore’s highly multilingual and rapidly developing sociolinguistic environment \citep{lee2016defense}.


Despite its cultural significance, Singlish has received limited attention from a computational perspective. In recent years, however, Singlish has become more visible in the written form, enabling systematic study of its evolution in everyday contexts \cite{gonzales2023corpus}. As the sociolect shifts, it creates a ``moving target''  for computational models. An LLM trained on historical data may fail to capture the nuanced stylistic trajectory of modern usage. This disconnect motivates two main research questions: \textbf{RQ1:} How has Singlish evolved stylistically over time? and \textbf{RQ2:} Can LLMs generate temporally neutral Singlish that captures the underlying essence of the variety rather than the surface style of a particular year? Our goal is not to test temporal invariance, but to test whether models can abstract away from era-specific markers to approximate the stable structural core of the variety.



To study this, we analyze Singlish text messages from 2012 to 2021 and compare yearly subsets using interpretable linguistic features and encoder-derived representations. We then evaluate whether LLM-generated text aligns with the temporal profile of the corpus or collapses it into a flattened style. This lets us examine both diachronic change in real usage and the extent to which current models can recover a temporally neutral Singlish baseline.


\paragraph{Our \textbf{contributions} are:}
\begin{itemize}
    \item \textbf{Stylistic similarity framework}: We propose an interpretable metric of stylistic divergence using both hand-crafted linguistic features and encoder-derived representations.
    \item \textbf{Diachronic analysis of Singlish}: We show systematic shifts that can be explained using linguistic and psycholinguistic features.
    \item \textbf{Evaluation of LLMs as Temporal Probes}: 
Using a null distribution baseline, we show that, at their most authentic prompting configuration, most current LLMs do not achieve authentic and temporally neutral Singlish simultaneously. Models generating realistic Singlish tend to inherit its temporal biases, while temporally neutral models tend to produce generic, non-Singlish-like outputs. 
\end{itemize}



\section{Related Work}

Research on Singlish has its foundations in sociolinguistics. Such work documents substrate influences from dialects of the neighboring regions and shows how Singlish diverges structurally from standard English \cite{ningsih2023exploring,tan2023curious}. Pragmatic studies show how discourse particles such as \textit{lah}, \textit{lor}, and \textit{meh}, can function as stance markers and encode solidarity and affective nuances between speakers \citep{wong2004particles,mian1993dynamics,foo2024disentangling}. These works establish Singlish as a linguistically rich, contact-driven variety, making it a compelling area of study.

In computational sociolinguistics, prior work has examined diachronic language change in online settings, identifying lexical, semantic, and stylistic shifts over time across platforms such as X and Reddit \citep{eisenstein2014diffusion,grieve2018mapping}. While these studies provide general frameworks for 
analyzing temporal variation, Singlish remains underexplored in this context. The availability of large-scale corpora such as the Corpus of Singapore English Messages (CoSEM) \citep{gonzales2023corpus} enables systematic analysis of Singlish evolution in digital communication.

Computational work on Singlish has primarily focused on handling its non-standard syntax through annotation, translation, and normalization pipelines \citep{liu2022singlish,chen2015corpus}, with downstream applications in sentiment analysis, opinion mining and topic detection on social media and e-commerce platforms \citep{bajpai2016developing,hsieh2022singlish}. More recently, LLMs have been used to model Singlish semantics and style \citep{periti2024lexical}, support translation and hate-speech detection \citep{ng2024talking,ng2024sghatecheck}, and construct lexical resources \citep{chow2024word}. Although prompting and fine-tuning have been explored to align LLM outputs with community-specific language use \citep{ng2025examining,wu2024semantic}, these approaches often struggle to reproduce authentic and temporally grounded variation \citep{yong2023prompting}. Our work extends this line of research by testing whether LLMs can generate Singlish that is not only realistic, but also temporally neutral across years.

\section{Measuring Stylistic Similarity}
To measure stylistic similarity, we represent texts as feature vectors and train a classifier to differentiate between texts from different time periods.

\subsection{Feature Extraction}

\paragraph{\textbf{Handcrafted Features}}
We extracted three categories of handcrafted features to capture stylistic variation associated with broader sociolinguistic trends in Singlish usage \citep{ziegeler2020changes,ningsih2023exploring}: \textbf{Lexico-structural attributes}  (length, repetition, sentence complexity), \textbf{Functional pragmatic markers} (discourse particles, emojis, emoticons), and \textbf{Psycholinguistic dimensions} via the LIWC-22 lexicon (LIWC). The LIWC covers aspects of affect, cognition, and social orientation \citep{tausczik2010psychological}, capturing shifts in valence and attitude. A full feature list is provided in Appendix~\ref{tab:features}.


    
    Lexico-structural attributes and functional pragmatic markers are computed at the message level. In contrast, LIWC features are extracted from messages aggregated by year. This aggregation is necessary as LIWC scores are unreliable for extremely short texts \citep{zhao2016evaluating}. Aggregating by year provides stable psycholinguistic estimates while preserving the temporal resolution required for diachronic analysis.

\paragraph{\textbf{Encoder-derived Features}}

Besides handcrafted features, we also use dense sentence embeddings. These features are dense vector representations of messages derived from a pre-trained all-MiniLM-L6-v2 sentence encoder\footnote{\url{https://huggingface.co/sentence-transformers/all-MiniLM-L6-v2}}.  The decision to utilize a non-Singlish-adapted model was a deliberate methodological choice. Fine-tuning the encoder on Singlish data would likely introduce specific linguistic biases, potentially compromising the model's ability to generalize across different temporal variations. These 384-dimensional embeddings capture rich semantic and contextual properties, leveraging large-scale pre-training to model the subtle stylistic and temporal shifts inherent in Singlish discourse.

\subsection{Machine Learning Classifier Differentiation}
\label{subsec:MLCD}

To measure stylistic similarity, we used a Gradient Boosting Classifier from the scikit-learn library\footnote{\url{https://scikit-learn.org/stable/modules/generated/sklearn.ensemble.GradientBoostingClassifier.html}} on feature vectors representing two sets of messages (e.g., 2012 vs. 2013) using an 80:20 random train:test split. To account for substantial variation in yearly message volume and ensure consistent statistical power across all year-pair comparisons, we fixed the sample size for each year in every pairwise comparison to n = 13,877, matching the smallest available yearly subset (2012), before the 80:20 train:test split. This uniform sampling budget across all comparisons prevents both class-frequency imbalance within a pair and inconsistent statistical power across gap sizes, ensuring that comparisons at large temporal gaps are not systematically noisier than those at small gaps due to differing sample sizes. We additionally evaluated Logistic Regression on the encoder-based features as a robustness check; the resulting temporal-gap trend was consistent with the Gradient Boosting Classifier results, as shown in Appendix~\ref{appendix:different_classifiers}. We ultimately selected the Gradient Boosting Classifier because its ensemble nature provides better stability and robustness to noise inherent in sociolectal data \citep{gope2024multi}. Furthermore, the choice of classifier was guided by preliminary experiments favoring simplicity and predictive power. To ensure robustness, training and testing were repeated three times per year-pair, reporting mean accuracy and standard deviation. 





\subsection{Similarity Metric}
We use classifier accuracy as a proxy for stylistic separability. In a balanced binary classification task, a classifier’s performance relative to the 50\% chance baseline indicates signal strength. When two year-subsets are stylistically similar, the classifier should perform near chance (\textit{acc}=50\%), whereas higher accuracy indicates stronger discriminability. We thus transform accuracy into a similarity score to obtain a monotonic, interpretable measure of cross-year stylistic distance. This metric is relative rather than absolute, intended to compare year pairs under a consistent evaluation setup.



To obtain an interpretable score, we transformed the raw accuracy (\textit{acc}) into a normalized similarity score, $S = 1 - 2 |\textit{acc} - 0.5 |$ where $0 \leq S \leq 1$.
Here, $S=1$ (\textit{acc} = 0.5) indicates maximal similarity and $S = 0$ (\textit{acc} = 0.0 or 1.0) indicates perfect discriminability. Accuracies below 50\% are rare in our data and, when they occur, lie close to chance; this reflects sampling noise rather than the classifier performing poorly, and such cases are treated identically to above-chance accuracies since only distance from chance is meaningful under our similarity construct.


By utilizing accuracy, we treat both temporal periods with equal weight, ensuring the metric reflects the global relationship between the two datasets. This allows for a consistent comparison of stylistic evolution across various time intervals and between human-authored and synthetic texts.

\begin{figure*}[ht!]
    \centering
    \includegraphics[width= 0.95\linewidth]{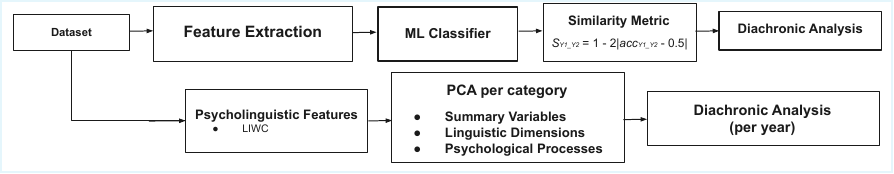}
    \caption{Pipeline for our diachronic analysis of CoSEM and LLM-generated Singlish texts.}
    \label{fig:pipeline}
\end{figure*}

\subsection{Temporal Neutrality and Realism Metrics}

To evaluate whether model outputs are biased towards particular time periods, we computed the standard deviation, $\sigma$, of similarity scores across years. Standard deviation captures temporal fluctuation in scores and provides a scale-independent summary comparable across models. Lower deviation indicates greater temporal neutrality, i.e., outputs that are less dependent on year-specific features and more representative of Singlish overall. We use this metric to assess whether prompting or fine-tuning reduces diachronic bias.

We evaluate realism by training classifiers to discriminate between real CoSEM messages and LLM generations where a higher similarity ($S$) denotes greater overall authenticity, while a low standard deviation ($\sigma$) across year-pairs signals temporal neutrality. This controls for the possibility 
that low variance reflects generic rather than genuine temporally neutral Singlish generation.



\section{Methodology}
Figure~\ref{fig:pipeline} illustrates our methodological pipeline.

\subsection{Dataset}
We used text messages sent from the years 2012 to 2021 from the \textbf{CoSEM} dataset \cite{gonzales2023corpus}, excluding 2022 due to substantially lower 
message volume. Table \ref{tab:per-year-counts} in Appendix \ref{appendix:counts} reports the number of CoSEM messages sampled per year. The smallest yearly subset (2012, n = 13,877) is still  much larger than the 100 messages generated per model per scheme (Section \ref{sec-4.3}), supporting the adequacy of this sampling scale relative to the real yearly volume.



\subsection{Content Impact Isolation}
\label{appendix:LDA}
To isolate style from topical shifts, we trained a latent Dirichlet allocation (LDA) model 
($K=15$) on yearly bins and measured cross-year variation via 
Jensen-Shannon divergence and cosine similarity \cite{hong2010}. Yearly topic distributions were highly stable with a mean pairwise JS divergence of 0.0178 ($\sigma^2 = 0.008$) and mean cosine 
similarity of 0.998 ($\sigma^2 = 0.001$), confirming that temporal effects in subsequent analyses reflect stylistic rather than topical change.

\subsection{Diachronic analysis of Real-World Data}

To examine diachronic changes in Singlish, we conducted pairwise classification experiments on year-pairs grouped by temporal gap. For instance, 2012 was paired with 2013, then with 2014, and so on. This design tests the hypothesis that linguistic differences accumulate over time, making temporally distant texts more distinguishable.

 Grouping by temporal gap, rather than comparing individual year pairs, allows us to isolate temporal effects from individual year-specific variation. Across ten years of data, this yielded 45 unordered year-pairs, spanning one- to nine-year gaps. For each year pair, we trained a classifier and aggregated results by temporal gap. Higher classification accuracy, or equivalently lower similarity, for larger gaps would indicate observable stylistic or structural shifts in Singlish over time, while flat or inconsistent trends would suggest relative linguistic stability.

\subsection{Generating Singlish Texts with LLMs} \label{sec-4.3}
We evaluate whether LLMs generate texts that are biased towards any particular time period and investigate the effects of fine-tuning on realism and temporal neutrality. We selected four pre-trained LLMs from HuggingFace with their default configurations: Qwen/Qwen-7B-Chat\footnote{\url{https://huggingface.co/Qwen/Qwen-7B-Chat}}, 
mistralai/Mistral-7B-Instruct-v0.1\footnote{\url{https://huggingface.co/mistralai/Mistral-7B-Instruct-v0.1}}, SeaLLMs/SeaLLM-7B-v2\footnote{\url{https://huggingface.co/SeaLLMs/SeaLLM-7B-v2}}, deepseek-ai/DeepSeek-R1-Distill-Qwen-1.5B\footnote{\url{https://huggingface.co/deepseek-ai/DeepSeek-R1-Distill-Qwen-1.5B}}, hereafter referred to as Qwen, Mistral, SeaLLM and DeepSeek-R1 respectively.

We performed supervised fine-tuning on Qwen and Mistral using LoRA \citep{LORA} within the LLaMAFactory framework, adopting default hyper-parameters \citep{zheng2024llamafactory}. The resulting models, Qwen-FT and Mistral-FT, were fine-tuned on the full CoSEM dataset (2012–2021). Training on the full pan-temporal distribution 
aims to suppress localized historical artifacts in favor of a hypothesized invariant Singlish core.


For text generation, each model was given a single-sentence Singlish dialogue seed, then prompted to continue the conversation, reflecting the conversational style of CoSEM messages. We kept the seed to a single sentence to balance two competing concerns: providing sufficient conversational context for a coherent continuation, while avoiding a seed long enough to strongly determine the model's stylistic output. We evaluated four prompting strategies (full prompts in Appendix \ref{appendix:prompts}):
\begin{itemize}
    \item \textbf{Zero-Shot (ZS): } Direct continuation in Singlish using default generation settings capturing the model's untreated Singlish style.
    \item \textbf{Chain-Of-Thought (CoT): } Explicit reasoning over cultural context and pragmatic intent prior to generation.
    \item \textbf{Diverse Decoding (DD): } Generation with increased stylistic and topical diversity.
    \item \textbf{Self-Consistency (SC): } Multiple independent continuations followed by self-selection of the most plausible output.
\end{itemize}

For each model and prompting scheme, we generated 100 messages, resulting in 2,400 generated texts.

\subsection{Diachronic Alignment of LLM Outputs with CoSEM Data}
We evaluated diachronic alignment by comparing LLM-generated texts against CoSEM messages from each year using the same stylistic similarity pipeline. Specifically, we computed similarity between each model’s outputs and yearly CoSEM subsets: (LLM vs 2012), (LLM vs 2013), …, (LLM vs 2021). Higher similarity to specific years indicates temporal anchoring, while stable scores suggest a more temporally neutral representation.


\subsection{Null Distribution for Temporal Neutrality}
A key limitation of variance-based temporal neutrality assessment is the absence of a principled baseline: any nonzero variance could in principle be flagged as temporal bias, even if it falls within the expected range of random sampling noise. Therefore, we constructed a null distribution of stylistic variance representing expected variance under the assumption of no temporal structure. The 95th percentile of this null variance distribution serves as our formal threshold $\sigma_{95}$. A model whose cross-year variance falls below $\sigma_{95}$ is statistically 
indistinguishable from temporally neutral texts. 

Concretely, we repeatedly sampled two groups of n = 13,877 messages each (matching the fixed sample size used in Section \ref{subsec:MLCD}) uniformly at random from the full CoSEM pool without respect to year labels, and computed the stylistic similarity between them using the classifier pipeline in Section \ref{subsec:MLCD}. Since both groups are drawn from the same undifferentiated pool, any observed variance in similarity scores across iterations reflects sampling noise alone rather than a genuine temporal signal. This procedure was repeated in $n_{outer} = 500$ outer iterations, each comprising $n_{inner} = 10$ pairwise comparisons, resulting in a distribution of 500 variance estimates under the null hypothesis of temporal neutrality.


\section{Results}

\subsection{Diachronic Differentiability as a Function of Temporal Gap}
\begin{figure}
    \centering
    \includegraphics[width=0.9\linewidth]{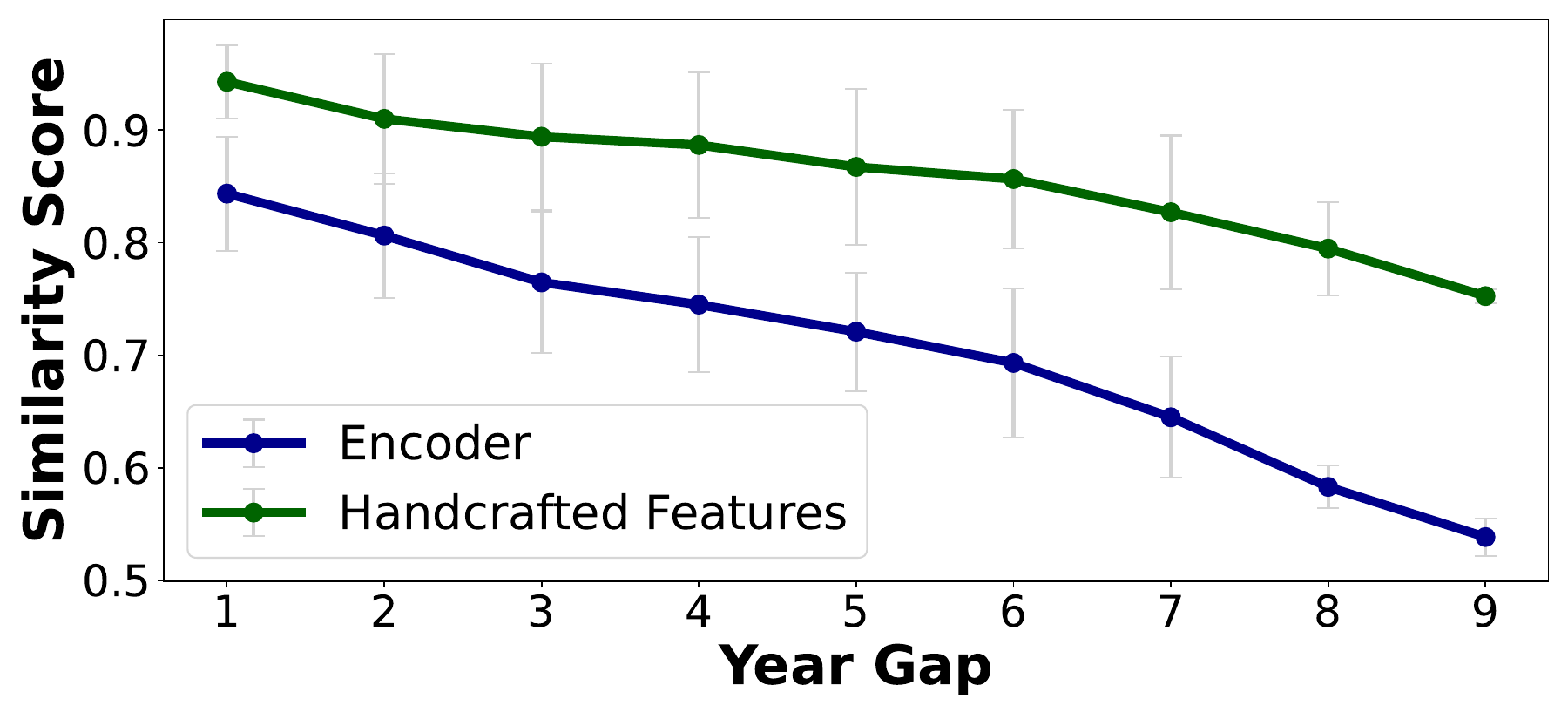}
    \caption{On CoSEM data, similarity score decreases as the year gap between data increases. }
    \label{fig:similarity_year_gap}
\end{figure}

Stylistic separability increases with temporal distance for both encoder-derived and handcrafted features. For encoder-derived representations, similarity decreases from 0.843 at T=1 to 0.538 at T=9. Handcrafted features show the same monotonic pattern, with similarity decreasing from 0.943 at T=1 to 0.754 at T=9 (see Figure \ref{fig:similarity_year_gap}). The Spearman correlation between the two trend lines is 1.00 (p < 0.05), indicating that the encoder has captured many of the same interpretable distributional properties explicitly modeled in the handcrafted feature set, while maintaining a different scaling.

\subsection{Qualitative Linguistic Shifts}
To ground these performance metrics in observable change, we highlight representative examples that mirror this increasing separability. Below are examples where individuals described their stress:

\begin{itemize}
\item \textbf{2013}: No I will stress until die one
\item \textbf{2014}: Walao liddat I stress sia
\item \textbf{2020}: I really nvr feel so stress about anything before
\end{itemize}

The 2013 and 2014 text messages appear to be more similar in style, as they exhibit more basilectal features characterized by the use of local loanwords and non-standard, substrate-influenced grammar. Meanwhile, that of 2020 appears to be more acrolectal or mesolectal in nature. This mirrors the similarity scores of 0.82 and 0.64 for 2013 vs 2014 and for 2013 vs 2020, respectively. Although these examples are illustrative rather than definitive, they help ground the temporal-gap results in observable linguistic change.

\subsection{Feature Importance} \label{sec-5.3}
\begin{figure*}[t]
    \centering
    \includegraphics[width=0.7\linewidth]{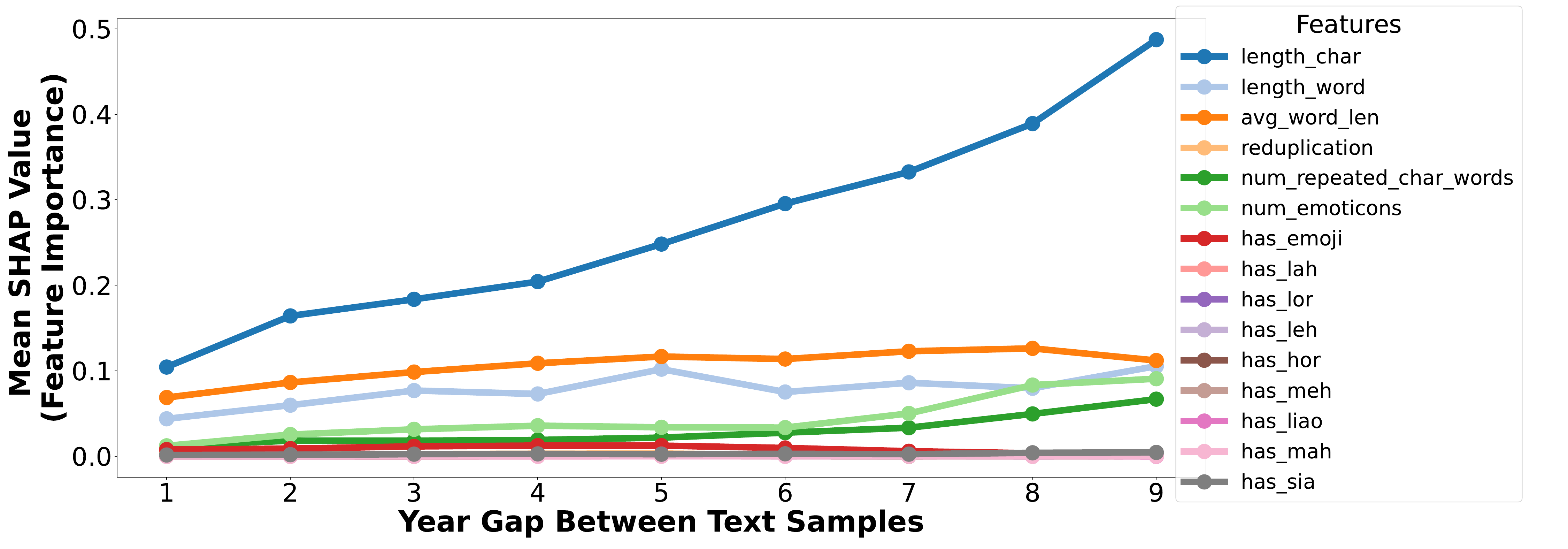}
    \caption{Feature Importance of each of the handcrafted features plotted against year-gap. The feature importance is measured by taking the mean (absolute) SHAP values.}
    \label{shap-values}
\end{figure*}

We calculated the mean absolute SHapley Additive exPlanations (SHAP) feature importance scores \citep{lundberg2017unified} of the handcrafted features to identify contributors to diachronic discriminability. The mean absolute SHAP values and statistical tests are reported in Appendix \ref{tab:shap_feature_stats}. 
Overall, from Figure \ref{shap-values}, we observe that most features have a steadily increasing importance across the decade of text messages, reflecting their cumulative role in capturing diachronic divergence. Among these, length-related features (i.e., length\_char, length\_word, avg\_word\_len) were the most influential, with character length exhibiting the steepest increase.

A second tier of features, including repeated-character words, Unicode emojis, ASCII emoticons, reduplication, and the particles \textit{leh} and \textit{lah}, displayed lower (mean |SHAP| $< 0.10$) but statistically significant contributions. In contrast, most other Singlish particles exhibited statistically non-significant trends over time, suggesting that though they function as stable identity markers of Singlish, they are weak predictors of temporal variation.



  

\begin{figure*}[htbp]
    \centering
    \begin{minipage}[t]{0.32\textwidth}
        \centering
        \includegraphics[width=\linewidth]{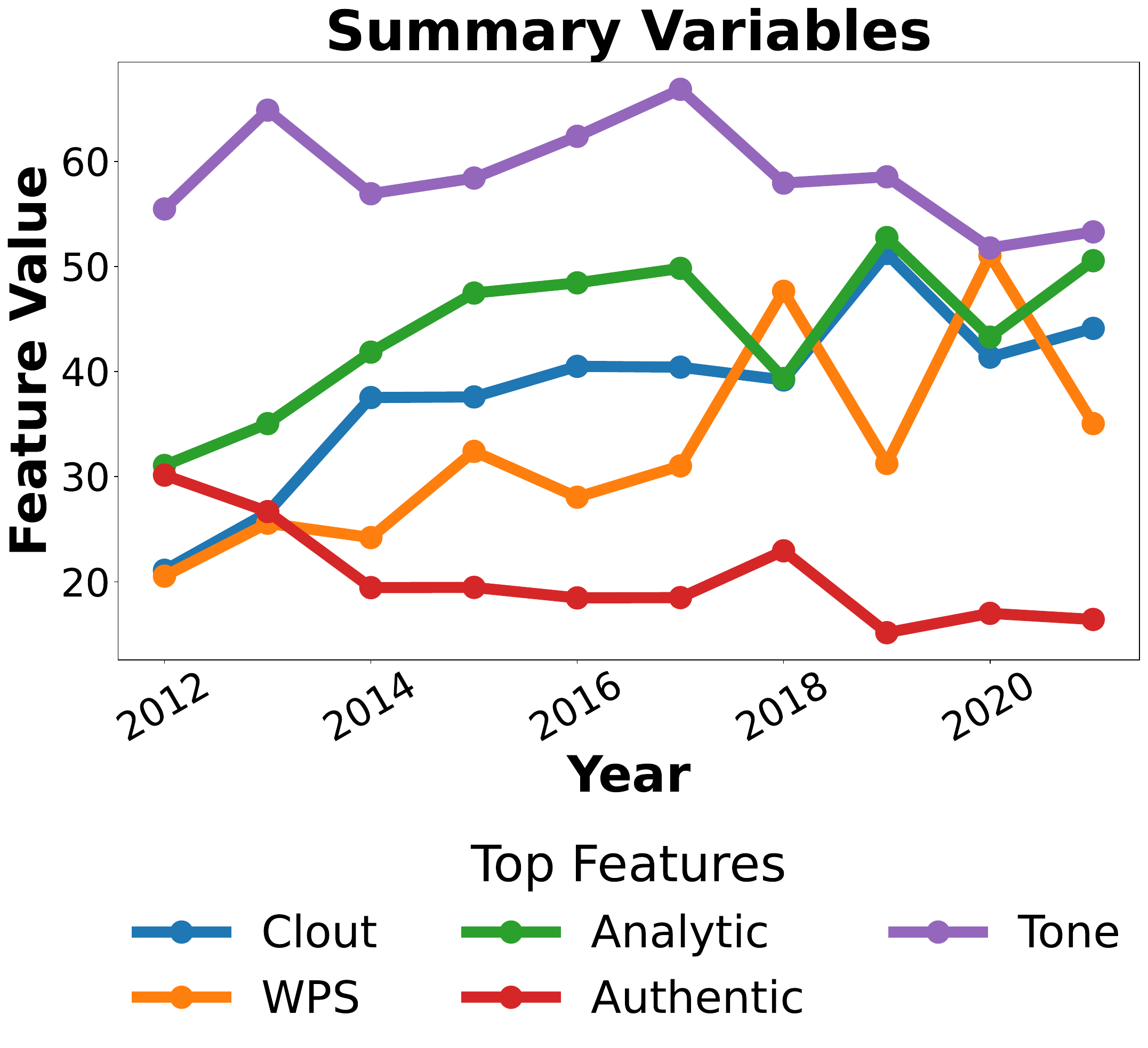}
        \\[2pt]
        \small (a) Summary Variables
    \end{minipage}
    \hfill
    \begin{minipage}[t]{0.32\textwidth}
        \centering
        \includegraphics[width=\linewidth]{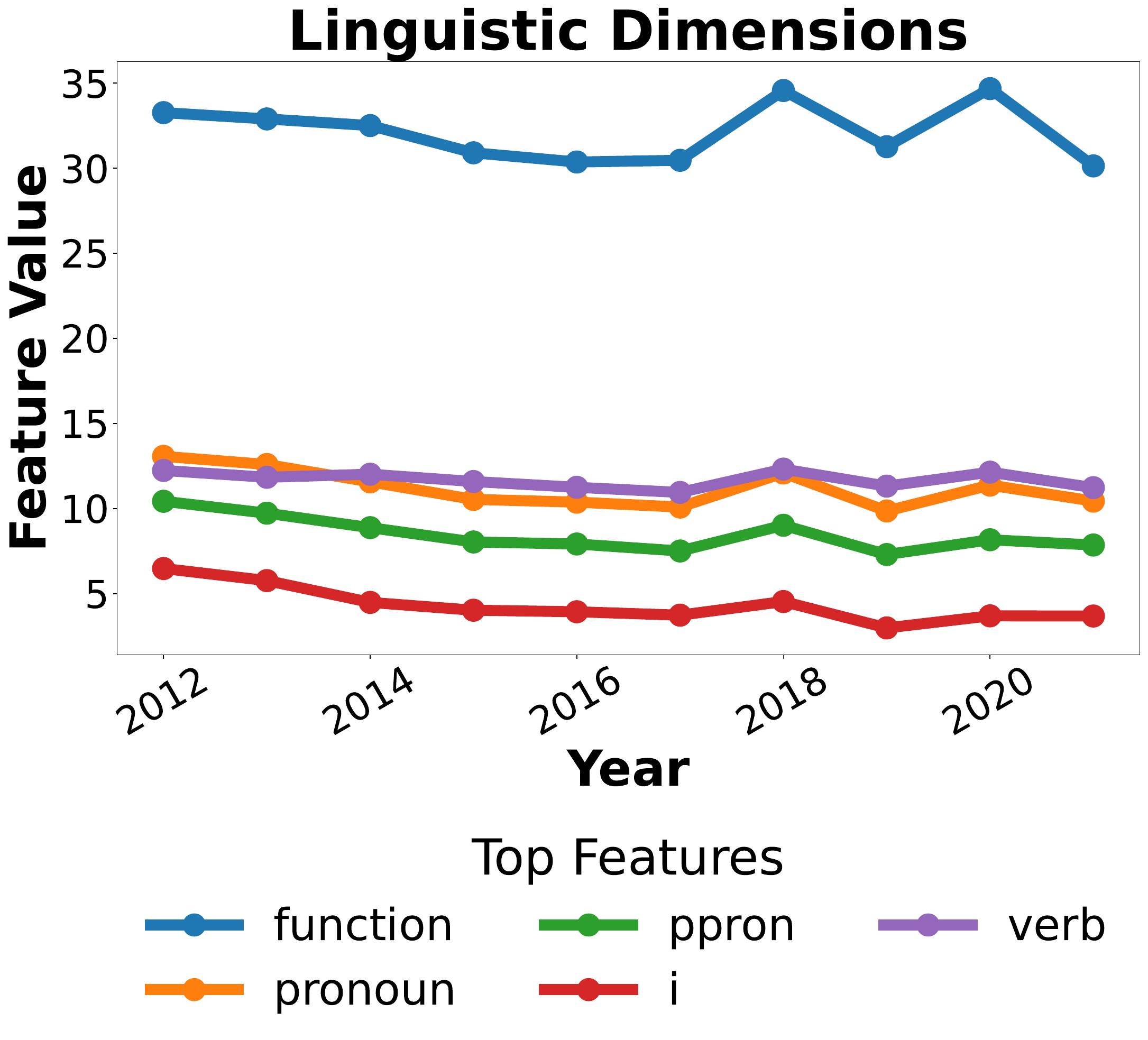}
        \\[2pt]
        \small (b) Linguistic Dimensions
    \end{minipage}
    \hfill
    \begin{minipage}[t]{0.32\textwidth}
        \centering
        \includegraphics[width=\linewidth]{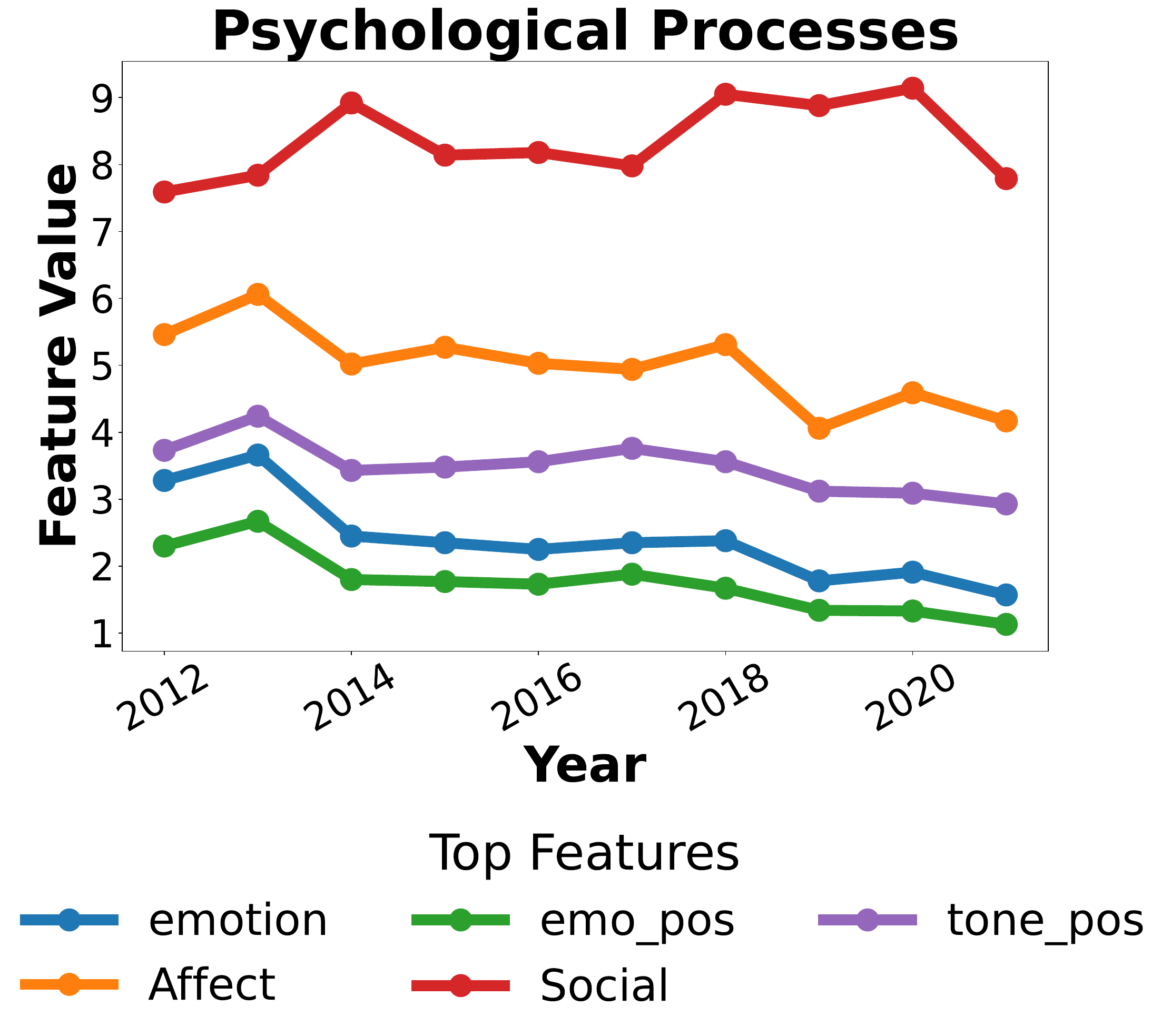}
        \\[2pt]
        \small (c) Psychological Processes
    \end{minipage}

    \caption{Trend of top LIWC features across categories.}
    \label{fig:liwc_top_features}
\end{figure*}



\subsection{Diachronic Shifts in LIWC Features}

To provide a more interpretable account of linguistic change, we applied Principal Component Analysis (PCA) to LIWC features for each year and tracked the five highest-loading features within each of the three LIWC categories. Temporal trends are shown in Figure \ref{fig:liwc_top_features} and the full statistical analyses are summarized in Appendix \ref{appendix:LIWC}.

\textbf{Summary variables} (Figure~\ref{fig:liwc_top_features}a) There is a significant increase in \textit{Clout}-, \textit{WPS}-, and \textit{Analytic}-related features, indicating shifts toward greater formality and structural complexity. By contrast, \textit{Tone} showed no consistent trend. This pattern indicates that Singlish texts in recent years adopt a more structured and polished form compared to earlier corpora while its tone varies across the years.

\textbf{Linguistic dimensions} (Figure~\ref{fig:liwc_top_features}b) Most features remained stable, except for first-person singular pronouns (\textit{i}) which declined over time. This reduction in self-referential markers could suggest a reduction in explicit personal discourse which we define as an individual’s attempt to make their presence overtly visible through expressions of personal involvement. We can also observe this trend qualitatively in messages from different years, each announcing the sender's arrival: 

\begin{description}
    \item[\textbf{2012:}] (a) \textit{Wait I'm reaching} \quad (b) \textit{I reaching my house in 10min like that}
    \item[\textbf{2021:}] (a) \textit{Reaching 9.20} \quad (b) \textit{Omw to nex}
\end{description}

In the 2012 samples, the subject (e.g., ``I'') is explicitly marked. By 2021, the messages drop the subject to focus squarely on the action. This observation is consistent with our quantitative findings showing a decline in self-referential markers alongside a shift toward structured and analytic discourse.



\textbf{Psychological processes} (Figure~\ref{fig:liwc_top_features}c) Affective expression shows a downward trend, with positive emotions and tone in particular decreasing across the decade. However, social process features remained stable, showing reduced emotional expressivity without diminished interpersonal orientation.

Collectively, LIWC analyses reveal that Singlish messages have become more structured and confident but less affective and self-focused over time.

\subsection{LLM Realism and Temporal Neutrality}

LLM performance varies substantially across architectures and 
prompting strategies. Among the base models, Mistral achieved 
the highest similarity (authenticity) under the DD prompting scheme ($S = 0.523$), 
followed by SeaLLM with CoT ($S = 0.417$), Qwen with CoT 
($S = 0.337$), and DeepSeek-R1 with SC ($S = 0.321$). 
Fine-tuning improves realism for some models and degrades it 
for others: Qwen-FT showed a substantial improvement ($S = 0.673$ 
with CoT), whereas Mistral-FT underperformed relative to its 
base counterpart ($S = 0.437$), indicating that fine-tuning does not uniformly help (see Table~\ref{tab:similarity_methods}).

Comparing each model's best-performing scheme 
against the null threshold $\sigma_{95} = 0.034$, DeepSeek-R1 ($\sigma = 0.028$) and SeaLLM ($\sigma = 0.033$) achieve temporal neutrality, while Mistral ($\sigma = 0.054$), Mistral-FT ($\sigma = 0.059$), Qwen ($\sigma = 0.043$), and Qwen-FT ($\sigma = 0.043$) retain detectable year-specific signals. A supplementary $\chi^2$ test (Appendix~\ref{sec_variance_test}) supports this split with varying strength: Mistral and Mistral-FT significantly exceed $\sigma_{95}$ ($\chi^2(9)=22.99$, $p=0.006$ and $\chi^2(9)=26.74$, $p=0.0015$, respectively), whereas Qwen and Qwen-FT exceed $\sigma_{95}$ numerically but do not reach conventional significance ($p=0.101$ and $p=0.104$), indicating weaker statistical evidence of temporal anchoring for these two models.

At each model's most authentic scheme, multi-year fine-tuning did not suppress temporal signals. Mistral-FT showed increased variability relative to its base counterpart ($\sigma$ = 0.059 vs. 0.054), while Qwen-FT's variability was unchanged ($\sigma$ = 0.043). Apart from SeaLLM, the only model below $\sigma_{95}$ at its best scheme, DeepSeek-R1, is also the least authentic, while the most authentic base model, Mistral, significantly exceeds the threshold. The trade-off is not universal across configurations. Qwen-FT under DD (S = 0.659, $\sigma$ = 0.021) and SC (S = 0.570, $\sigma$ = 0.031) falls below $\sigma_{95}$ while remaining the most authentic model (Table 1), indicating that fine-tuning combined with some prompting strategies can yield both properties.

\subsection{Qualitative Error Analysis}\label{sec:qualitative}

To better interpret shifts in realism, we examined 
model outputs for cultural and grammatical nuances, 
focusing on what distinguishes models across the 
authenticity-neutrality spectrum.

DeepSeek-R1, the most temporally neutral model ($\sigma = 0.028$) but least authentic ($S = 0.321$), produced outputs notably lacking Singlish-specific features, such as \textit{``Okay, so you're going to go out later.''} -- using standard English with no discourse particles, code-switching, or colloquial markers. This suggests that its temporal 
neutrality reflects an absence of sociolectal grounding rather than genuine abstraction of invariant Singlish structure.

SeaLLM presents a more nuanced case: despite achieving temporal neutrality ($\sigma = 0.033$), it attained the second highest authenticity among base models ($S = 0.417$), producing outputs that incorporated Singlish particles more consistently than DeepSeek-R1, such as \textit{``Yo, I'm thinking of it lah. What you up to?''}.

Among the more authentic but temporally biased models, outputs varied in depth of Singlish grounding. Qwen-FT struggled with phonetic authenticity despite higher similarity scores (more authentic), generating \textit{``Naa 
lah. I'm just gonna stay at home and chill.''} The use of \textit{Naa} resembles an Americanized 
negation pattern; a culturally grounded speaker 
would use \textit{No lah}, revealing a gap in deep colloquial knowledge. 

%


Overall, four patterns emerge: (i) temporal neutrality most commonly reflects generic generation rather than abstracted Singlish, though SeaLLM's Southeast Asia-focused pretraining suggests broader exposure can partially decouple the two; (ii) the authenticity-neutrality trade-off 
holds strongly for most models at their most authentic scheme but is not absolute; (iii) prompting effectiveness and fine-tuning gains 
are model-dependent; and (iv) residual diachronic signals largely persist across high-authenticity models, indicating they remain partially temporally anchored rather than fully timeless.

\begin{table*}[ht]
\centering
\caption{Comparison of mean similarity scores and standard deviations across models and prompting schemes. Bold text indicates the highest mean similarity score for each model, denoting peak authenticity.}
\label{tab:similarity_methods}
\begin{tabular}{lcccccccc}
\hline
Model &
\multicolumn{2}{c}{Baseline} &
\multicolumn{2}{c}{CoT} &
\multicolumn{2}{c}{Diverse-Decoding} &
\multicolumn{2}{c}{Self-Consistency} \\
\hline
 & Mean & Std & Mean & Std & Mean & Std & Mean & Std \\
\hline
Mistral      & 0.367 & 0.024 & 0.444 & 0.056 & \textbf{0.523} & 0.054 & 0.373 & 0.043 \\
Mistral-FT   & \textbf{0.437} & 0.059 & 0.303 & 0.042 & 0.283 & 0.040 & 0.329 & 0.050 \\
Qwen         & 0.168 & 0.020 & \textbf{0.337} & 0.043 & 0.105 & 0.010 & 0.240 & 0.031 \\
Qwen-FT      & 0.668 & 0.034 & \textbf{0.673} & 0.043 & 0.659 & 0.021 & 0.570 & 0.031 \\
DeepSeek-R1  & 0.198 & 0.017 & 0.228 & 0.041 & 0.268 & 0.028 & \textbf{0.321} & 0.028 \\
SeaLLM       & 0.214 & 0.019 & \textbf{0.417} & 0.033 & 0.350 & 0.042 & 0.360 & 0.043 \\
\hline
\end{tabular}
\end{table*}


\section{Discussion}

\subsection{Singlish Particle Analysis}
\begin{figure}
    \centering
    \includegraphics[width=1\linewidth]{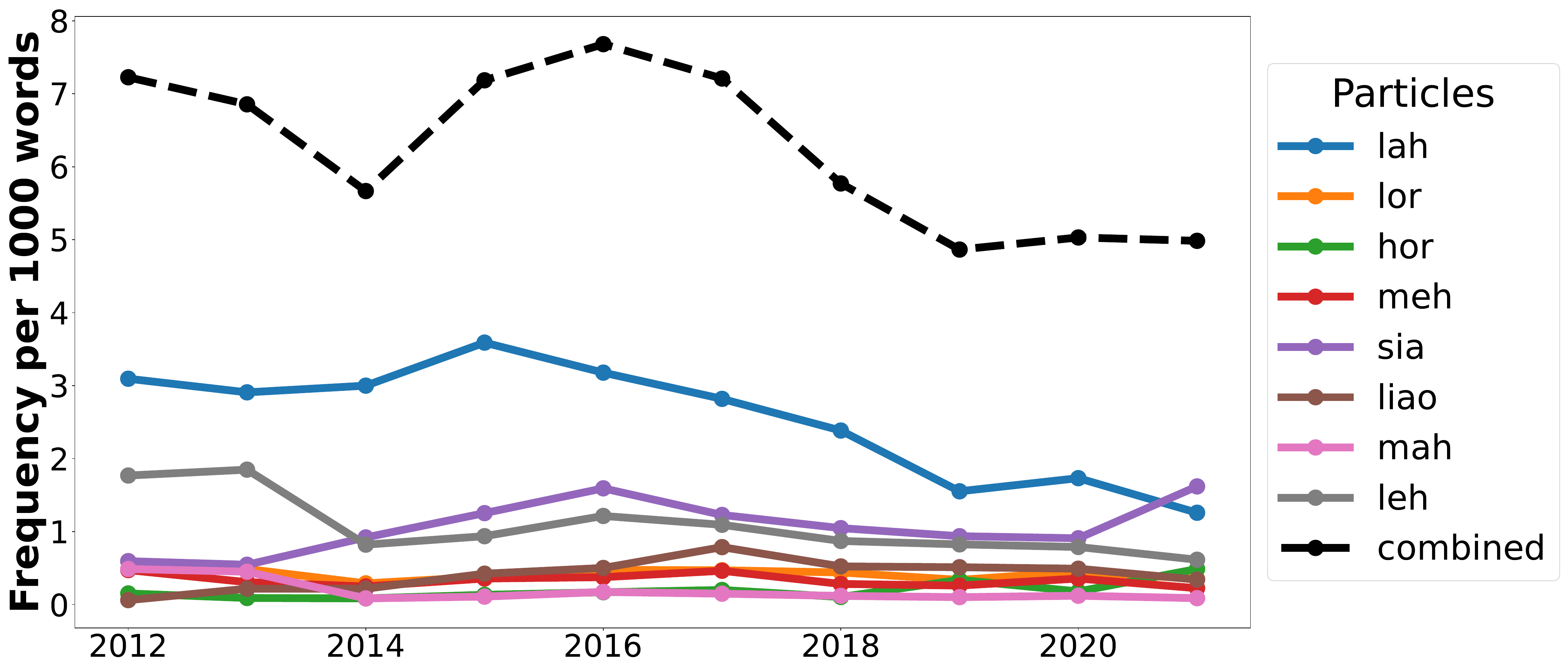}
    \caption{Normalized frequency/1000 words of individual (solid) and combined (dotted) Singlish particles in CoSEM over time.}

    \label{particle usage}
\end{figure}

SHAP analysis reveals that most Singlish discourse particles (\textit{mah}, \textit{liao}, \textit{lor}, \textit{hor}, \textit{meh}) contribute minimally to diachronic classification, while only high-frequency particles (\textit{lah}, \textit{leh}, \textit{sia}) show consistent model influence. Complementary LIWC analysis indicates a broader shift toward regularized, formal English structure, matching an overall decline in particle frequency since 2016 (see Figure \ref{particle usage}). 

Collectively, these patterns suggest that presence- and frequency-based metrics risk underestimating the structural role of particles, underscoring the need for interpretive frameworks that capture their layered pragmatic meanings.


\subsection{Selective Formalization and Decreolization}
Complementing the SHAP and encoder-derived analyses, key LIWC features provide interpretable grounding for these diachronic trends (see Figure \ref{fig:liwc_top_features}). Notable increases in \textit{Clout}, \textit{Analytic}, and words per sentence (\textit{WPS}), alongside declines in \textit{Authentic} markers and first-person pronouns, signal a structural shift toward more formalized, depersonalized, and audience-oriented discourse. Conversely, stability in social process categories indicates that Singlish preserves its primary sociolinguistic function as a medium for interpersonal engagement.

These empirical patterns align with the processes of selective formalization and partial decreolization along the post-creole continuum \citep{Sato_1994}. Messages from earlier years exhibit highly compressed, basilectal constructions (e.g., \textit{``Can one la''} from CoSEM 2012), whereas later texts adopt mesolectal, Standard English-aligned syntactic frameworks without abandoning distinct sociolectal identity markers (e.g., \textit{``u can do it one''} from CoSEM 2020). This transition from structural shortcuts to analytical complexity suggests that Singlish is a moving target. 


This adaptation toward broader structural intelligibility within borderless digital environments \citep{Singaporean_internet_chit_chat} underscores how Singlish dynamically scales its accessibility while retaining core identity-bearing features.


\subsection{Why LLMs Might Fall Short}
LLM outputs show limited authenticity for 
Singlish, with models' response to fine-tuning 
and prompting varying by architecture \citep{alizadeh2025open}. Critically, our null distribution analysis reveals that temporal neutrality does not imply success: models achieving neutrality (e.g., DeepSeek-R1) do so largely through generic rather than authentic Singlish generation, while the most authentic base model (Mistral) retains significant temporal bias. At each model's most authentic configuration, this inverse relationship points to a trade-off that models that learn authentic Singlish features inherit the temporal distribution of those features from historically situated training data, while models that avoid temporal anchoring do so by failing to acquire Singlish-specific structure in the first place. 


SeaLLM is an exception. It falls below $\sigma_{95}$ while producing relatively authentic outputs, suggesting that Southeast Asia-focused pretraining \citep{nguyen-etal-2024-seallms} may provide broader cross-temporal coverage of Singlish features. Qwen-FT under DD and SC is a second exception. These cases suggest that temporal neutrality is not always a symptom of generic generation, and that fine-tuning combined with certain prompts can partially decouple authenticity from temporal anchoring. 

Two factors likely explain why this trade-off persists. First, standard language model training optimizes next-token prediction based on distributional frequency \citep{mccoy-etal-2019-right} rather than abstract grammatical generalization. This reinforces epochal markers in 
the training data rather than filtering them out. At their most authentic schemes, our pan-temporal fine-tuned models suggest that rather than distilling a temporally invariant Singlish representation, Qwen-FT and Mistral-FT preserved transient historical variations, with Mistral-FT showing increased temporal variability relative to its base counterpart. Second, some variation may be irreducibly time-bound. Singlish operates along a post-creole continuum \citep{Sato_1994} where certain lexical and pragmatic markers are tied to changing social contexts, so stripping temporal context risks undermining the very features that make generated text recognizably Singlish.

Together, these factors suggest that temporal 
neutrality is a representational challenge rather than a prompting or fine-tuning one, requiring either temporally balanced training data or objectives that explicitly disentangle stable sociolectal structure from transient markers. These findings may extend beyond Singlish. For any other fast-evolving 
contact language or sociolect, LLMs are likely to inherit the temporal biases of their training data, and temporal neutrality as defined here serves as a diagnostic for whether a model has learned the structural core of a variety or merely a historical snapshot.

\section{Conclusion}
We examined stylistic shifts in Singlish across a decade of digital communication using a framework combining neural encoders, handcrafted features, and psycholinguistic dimensions. Singlish exhibits substantial diachronic divergence driven primarily by structural features, while most discourse particles are weak predictors of temporal change despite a decline in overall particle frequency. Aligning encoder representations with interpretable linguistic features and psycholinguistic dimensions reveals that much of the temporal separability learned by neural models can be explained through surface-level stylistic cues.

Against a null distribution baseline, LLMs largely exhibit a trade-off: models generating authentic Singlish tend to inherit its temporal biases, while temporally neutral models typically produce generic English outputs. SeaLLM, however, suggests that culture-specific pretraining can soften this trade-off. This positions our framework as a practical diagnostic tool for studying sociolectal grounding and temporal bias in LLMs across fast-evolving language varieties.

\section{Limitations} This study focuses on SMS-style messages and may not generalize to social media text, where interactional norms and stylistic conventions differ. Future work should extend the analysis to text-based social media data to examine whether similar diachronic and stylistic patterns persist.

Because character/word length is the dominant SHAP feature (Section \ref{sec-5.3}), we cannot fully rule out that part of the diachronic signal reflects general compression effects in Singlish's substrate languages (e.g., Chinese, Malay) rather than sociolectal change specific to the CoSEM period. Disentangling this would require cross-linguistic length-normalization baselines, which we leave to future work.

Future work could systematically ablate the seed-construction strategy such as varying seed length, or truncating a real message and asking the model to complete it. This could isolate the effect of prompting design from model-intrinsic stylistic bias.

\section{Ethical Considerations} The CoSEM dataset is publicly available and anonymized, containing no personally identifiable information. All analyses were conducted at the aggregate level, with a focus on linguistic patterns rather than individual users.

\bibliography{custom}

\appendix

\section{Number of messages per year}
Table \ref{tab:per-year-counts} shows the breakdown of the number of messages per year in the CoSEM dataset.
\label{appendix:counts}
\begin{table}[h]
\centering
\caption{Number of CoSEM messages per year used in the diachronic analysis. 2022 is excluded from all analyses due to its low volume.}
\begin{tabular}{lr}
\toprule
Year & \# Messages \\
\midrule
2012  &   13{,}877 \\
2013  &   42{,}792 \\
2014  &   77{,}921 \\
2015  &  210{,}354 \\
2016  &  346{,}540 \\
2017  &  265{,}285 \\
2018  &  177{,}496 \\
2019  &   90{,}446 \\
2020  &  137{,}770 \\
2021  &   20{,}887 \\
2022  &     131 \\
\bottomrule
\end{tabular}
\label{tab:per-year-counts}
\end{table}

    \section{Different Classifiers}
\label{appendix:different_classifiers}
\begin{figure}[H]
    \centering
    \includegraphics[width=1\linewidth]{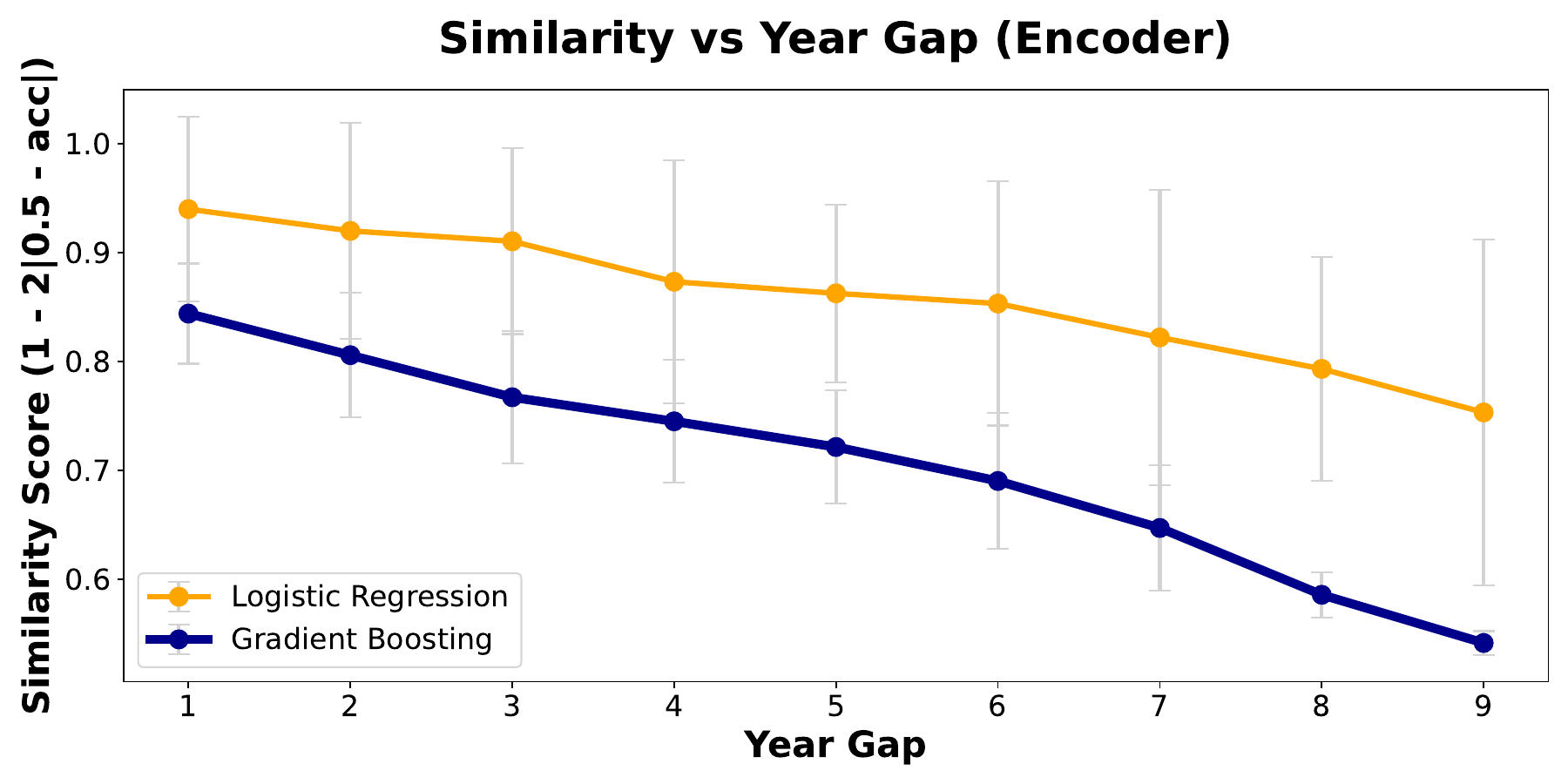}
    \caption{Similarity trends are consistent across classifiers.}
    \label{fig:placeholder}
\end{figure}


\section{Statistical Analysis of LIWC Features}
\label{appendix:LIWC}
See Table \ref{table-LIWC} for the statistical analysis for the LIWC features.

\begin{table}[ht]
\centering
\caption{Trend analysis of LIWC features over time. * are features with statistically significant trends ($p < 0.05$, Spearman correlation)}
\label{tab:liwc_trends_spearman}
\begin{tabular}{lccc}
\hline
\textbf{Feature} & \textbf{Slope} & \textbf{Spearman $\rho$} & \textbf{p-value} \\
\hline
\multicolumn{4}{l}{\textit{Summary Variables}} \\
Clout*     & 2.33  & 0.915 & $<$0.001 \\
WPS*       & 2.38  & 0.842 & 0.002 \\
Analytic*  & 1.60  & 0.697 & 0.025 \\
Authentic* & -1.23 & -0.818 & 0.004 \\
Tone               & -0.61 & -0.309 & 0.385 \\
\hline
\multicolumn{4}{l}{\textit{Linguistic Dimensions}} \\
function           & -0.07 & -0.188 & 0.603 \\
pronoun            & -0.22 & -0.588 & 0.074 \\
ppron*     & -0.24 & -0.648 & 0.043 \\
i*         & -0.28 & -0.842 & 0.002 \\
verb               & -0.05 & -0.309 & 0.385 \\
\hline
\multicolumn{4}{l}{\textit{Psychological Processes}} \\
emotion*   & -0.19 & -0.881 & 0.001 \\
Affect*    & -0.16 & -0.794 & 0.006 \\
emo\_pos*  & -0.14 & -0.915 & $<$0.001 \\
tone\_pos* & -0.10 & -0.711 & 0.021 \\
Social             & 0.08  & 0.382 & 0.276 \\
\hline
\end{tabular}
\label{table-LIWC}
\end{table}

\section{Different Prompting Schemes used to generate Singlish text}
\label{appendix:prompts}
The four prompts used to generate Singlish text are shown below. All use the same opening turn as the seed.

\noindent
\fbox{%
\parbox{\dimexpr\columnwidth-2\fboxsep-2\fboxrule\relax}{%
\ttfamily
\textbf{\underline{Zero-Shot}}\\
Generate a conversation in Singlish, with the starting text given below.\\
Only give the conversation as the output. Do not include anything else. The conversation can be as long as you want.\\ \\
Starting text: \\
\textbf{Person A}: Eh bro, you going out later or not?\\
\textbf{Person B}:
}}

\noindent
\fbox{%
\parbox{\columnwidth}{%
\ttfamily
\textbf{\underline{Chain-Of-Thought}}\\
Generate a conversation in Singlish, starting with the text below.\\
As you continue the conversation, for every turn, think about what a typical Singaporean might say in response, considering local culture, humour, and context. Do not include your reasoning—only output the conversation.\\ \\
Starting text:\\
\textbf{Person A}: Eh bro, you going out later or not?\\
\textbf{Person B}:
}}

\noindent
\fbox{%
\parbox{\columnwidth}{%
\ttfamily
\textbf{\underline{Diverse Decoding}} \newline
    Generate a conversation in Singlish, starting with the text below. \newline
    Aim for variety in topics - let the conversation flow naturally, touching on everyday Singaporean life (food, family, weather, work, etc). Your output should be only the conversation.\newline \newline
    Starting text:\newline 
    \textbf{Person A}: Eh bro, you going out later or not?\newline
    \textbf{Person B}:
    }}

\noindent
\fbox{%
\parbox{\columnwidth}{%
\ttfamily
\textbf{\underline{Self-Consistency}}\\
Generate a conversation in Singlish using the starting text below.\\
After each line, consider at least three plausible directions the conversation might take, and pick one that feels the most natural or interesting. Only output the conversation.\\ \\
Starting text:\\ 
\textbf{Person A}: Eh bro, you going out later or not?\\
\textbf{Person B}:
}}

\section{Variance Test Results}
\label{sec_variance_test}
Table \ref{tab_variance_test} presents the variance test results for the different LLMs and indicates the best-performing prompting methods.

\begin{table}[ht]
\centering
\caption{Variance test results for different models and best-performing prompting methods. $\chi^2$ tests whether cross-year standard deviation exceeds $\sigma_{95}=0.034$ (one-sided, $df=9$).}
\label{tab_variance_test}
\resizebox{\columnwidth}{!}{%
\begin{tabular}{l l c c c}
\hline
\textbf{Model} & \textbf{Method} & \textbf{Std} & \textbf{$\chi^2$ stat} & \textbf{p-val} \\
\hline
SeaLLM       & CoT      & 0.0330 & 8.47  & 0.4877 \\
DeepSeek-R1  & SC       & 0.0282 & 6.19  & 0.7212 \\
Mistral      & DD       & 0.0543 & 22.99 & 0.0062 \\
Mistral-FT   & Baseline & 0.0586 & 26.74 & 0.0015 \\
Qwen-FT      & CoT      & 0.0432 & 14.54 & 0.1045 \\
Qwen         & CoT      & 0.0434 & 14.64 & 0.1013 \\
\hline
\end{tabular}%
}
\end{table}

\section{Handcrafted Features}
\label{tab:features}
Table \ref{handcrafted_features} presents the details of the handcrafted features used in the stylistic similarity analysis.

\begin{table*}[t]
\centering
\caption{Features used in the stylistic similarity analysis.}
\label{handcrafted_features}

\begin{tabular}{p{7.5cm} p{5cm} p{1.6cm}}
\toprule
\textbf{Feature} & \textbf{Abbreviation} & \textbf{Unit} \\
\midrule

\multicolumn{3}{l}{\textbf{Handcrafted Features}} \\

\multicolumn{3}{l}{\textit{Lexico-structural attributes}} \\
No.\ of characters & length\_char & Message \\
No.\ of words & length\_word & Message \\
Average word length & avg\_word\_len & Message \\
No.\ of repeated characters (e.g., \textit{soooo}) & num\_repeated\_char\_words & Message \\
No.\ of repeated words (e.g., \textit{wait wait}) & reduplication & Message \\

\multicolumn{3}{l}{\textit{Functional pragmatic markers}} \\
Presence of discourse particles (e.g., \textit{lah, lor, leh}) & has\_\{particle\} & Message \\
Presence of Unicode emojis & has\_emoji & Message \\
No.\ of ASCII emoticons (e.g., \texttt{:P}) & num\_emoticons & Message \\

\midrule
\multicolumn{3}{l}{\textbf{Psycholinguistic Dimensions (LIWC; Tausczik and Pennebaker, 2010)}} \\

\textit{Summary variables} & & Year \\
Social confidence, words per sentence, logical thinking, authenticity, emotional tone
& Clout, WPS, Analytic, Authentic, Tone & \\

\textit{Psychological processes} & & Year \\
Emotion-related and social process words
& emotion, affect, emo\_pos, social, tone\_pos & \\

\textit{Linguistic dimensions} & & Year \\
Function words, pronouns, first-person pronouns, verbs
& function, pronoun, ppron, i, verb & \\

\midrule
\multicolumn{3}{l}{\textbf{Encoder-derived Features}} \\
Sentence-level vector representations from a pre-trained encoder & -- & Message \\

\bottomrule
\end{tabular}
\end{table*}



\section{Feature Importance for Diachronic Discriminability}
\label{tab:shap_feature_stats}
Table \ref{shap_feature_table} presents the results of the statistical analysis of SHAP feature importance.

\begin{table*}[ht!]
\centering
\caption{Statistical analysis of SHAP feature importance. * are features with Mean absolute SHAP values significantly above 0 ($p<0.05$)}
\label{shap_feature_table}
\begin{tabular}{lcccc}
\hline
Feature & Mean SHAP & Std SHAP & t-stat & p-value \\
\hline
length\_char*                & 2.68e-01 & 1.21e-01 & 1.46e+01 & 3.73e-07 \\
avg\_word\_len*              & 1.06e-01 & 1.85e-02 & 2.13e+01 & 4.83e-05 \\
length\_word*                & 7.80e-02 & 1.91e-02 & 9.75e+00 & 2.70e-04 \\
num\_emoticons*              & 4.42e-02 & 2.63e-02 & 6.51e+00 & 2.51e-04 \\
num\_repeated\_char\_words*  & 2.97e-02 & 1.78e-02 & 6.70e+00 & 2.24e-04 \\
has\_emoji*                  & 8.49e-03 & 3.62e-03 & 6.44e+00 & 1.28e-03 \\
has\_sia*                    & 2.89e-03 & 9.66e-04 & 4.50e+00 & 2.68e-03 \\
reduplication*               & 2.48e-03 & 6.47e-04 & 4.98e+00 & 8.61e-03 \\
has\_leh*                    & 1.68e-03 & 7.38e-04 & 3.92e+00 & 3.62e-03 \\
has\_lah*                    & 1.56e-03 & 5.73e-04 & 3.86e+00 & 4.50e-03 \\
has\_liao                   & 1.37e-03 & 9.55e-04 & 2.48e+00 & 6.01e-02 \\
has\_hor                    & 5.41e-04 & 8.71e-04 & 1.61e+00 & 1.48e-01 \\
has\_lor                    & 3.49e-04 & 1.38e-04 & 2.40e+00 & 5.54e-02 \\
has\_meh                    & 2.14e-04 & 1.66e-04 & 1.94e+00 & 1.00e-01 \\
has\_mah                    & 7.59e-05 & 7.13e-05 & 1.20e+00 & 2.59e-01 \\
\hline
\end{tabular}
\end{table*}
\end{document}